\documentclass[10pt,twocolumn,letterpaper]{article}

\usepackage[pagenumbers]{cvpr} 
 \usepackage{multirow} 

\usepackage{amsthm}
\theoremstyle{remark}
\newtheorem{remark}{Remark}
\definecolor{cvprblue}{rgb}{0.21,0.49,0.74}
\usepackage[pagebackref,breaklinks,colorlinks,allcolors=cvprblue]{hyperref}

\def\paperID{18202} 
\def\confName{CVPR}
\def\confYear{2026}

\title{MMT-ARD: Multimodal Multi-Teacher Adversarial Distillation \\ for Robust Vision-Language Models}

\author{
Yuqi Li\textsuperscript{1}\thanks{Equal Contribution.} \qquad
Junhao Dong\textsuperscript{2}\footnotemark[1] \qquad
Chuanguang Yang\textsuperscript{2} \qquad
Shiping Wen\textsuperscript{4}\\
Piotr Koniusz\textsuperscript{5} \qquad
Tingwen Huang\textsuperscript{6} \qquad
Yingli Tian\textsuperscript{1}\thanks{Corresponding Author.} \qquad
Yew-Soon Ong\textsuperscript{2}\footnotemark[2]
\\
\textsuperscript{1}The City University of New York, CUNY\\
\textsuperscript{2}Nanyang Technological University\\
\textsuperscript{3}Institute of Computing Technology, Chinese Academy of Sciences \\
\textsuperscript{4}University of Technology Sydney\\
\textsuperscript{5}Data61, CSIRO\\
\textsuperscript{6}Shenzhen University of Advanced Technology
}

\begin{document}
\maketitle
\begin{abstract}

Vision-Language Models (VLMs) are increasingly deployed in safety-critical applications, making their adversarial robustness a crucial concern. While adversarial knowledge distillation has shown promise in transferring robustness from teacher to student models, traditional single-teacher approaches suffer from limited knowledge diversity, slow convergence, and difficulty in balancing robustness and accuracy. To address these challenges, we propose MMT-ARD: a Multimodal Multi-Teacher Adversarial Robust Distillation framework. Our key innovation is a dual-teacher knowledge fusion architecture that collaboratively optimizes clean feature preservation and robust feature enhancement. To better handle challenging adversarial examples, we introduce a dynamic weight allocation strategy based on teacher confidence, enabling adaptive focus on harder samples. Moreover, to mitigate bias among teachers, we design an adaptive sigmoid-based weighting function that balances the strength of knowledge transfer across modalities. Extensive experiments on ImageNet and zero-shot benchmarks demonstrate that MMT-ARD improves robust accuracy by +4.32$\%$ and zero-shot accuracy by +3.5$\%$ on the ViT-B-32 model, while achieving a 2.3× increase in training efficiency over traditional single-teacher methods. These results highlight the effectiveness and scalability of MMT-ARD in enhancing the adversarial robustness of multimodal large models. Our codes are available at \url{https://github.com/itsnotacie/MMT-ARD}
\end{abstract}    
\section{Introduction}

With the rapid advancement of multimodal artificial intelligence technology, Vision-Language Models (VLMs) have been widely adopted in autonomous driving, medical imaging, and industrial inspection. By jointly learning visual and textual representations, these models demonstrate strong cross-modal reasoning abilities. However, VLMs remain highly vulnerable to adversarial perturbations. Studies show that adding imperceptible perturbations can lead to completely erroneous model predictions~\cite{016} such as traffic-sign misclassification in autonomous driving or diagnostic errors in medical settings. This fragility stems from the multimodal alignment mechanism of VLMs—attackers disrupt cross-modal attention calculations by perturbing critical regions in the visual feature space, causing the model to produce high-confidence erroneous matches for adversarial examples. As VLMs enter safety-critical applications, their adversarial vulnerability has emerged as a major security threat hindering technological deployment.
\begin{figure}[htbp]
    \centering
    \begin{subfigure}[b]{0.236\textwidth} 
        \includegraphics[width=\textwidth]{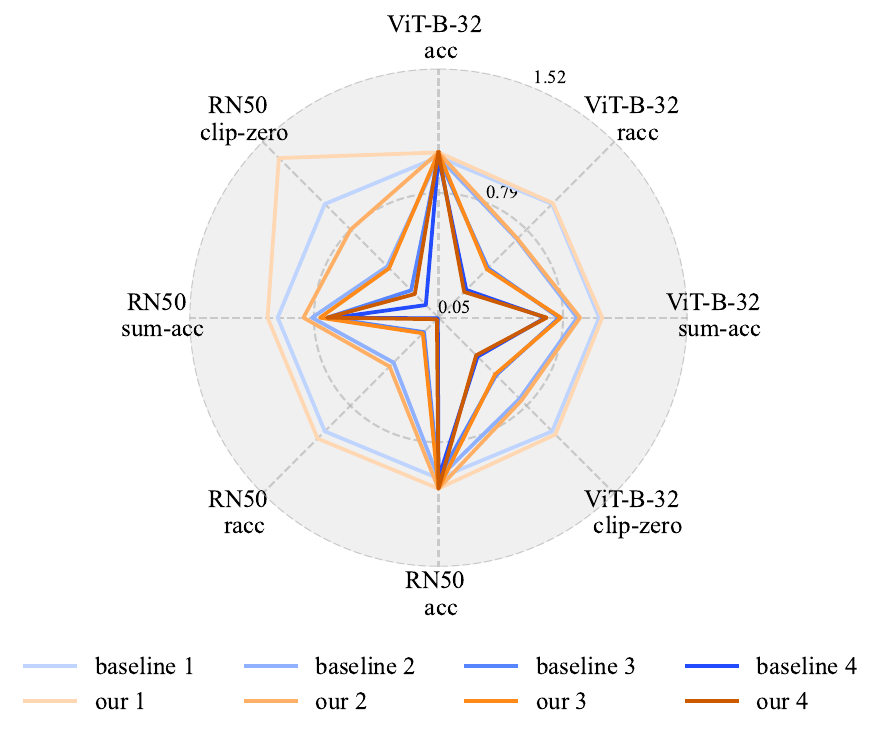}
        \caption{}
        \label{fig:sub1}
    \end{subfigure}
    \hfill 
    \begin{subfigure}[b]{0.236\textwidth}
        \includegraphics[width=\textwidth]{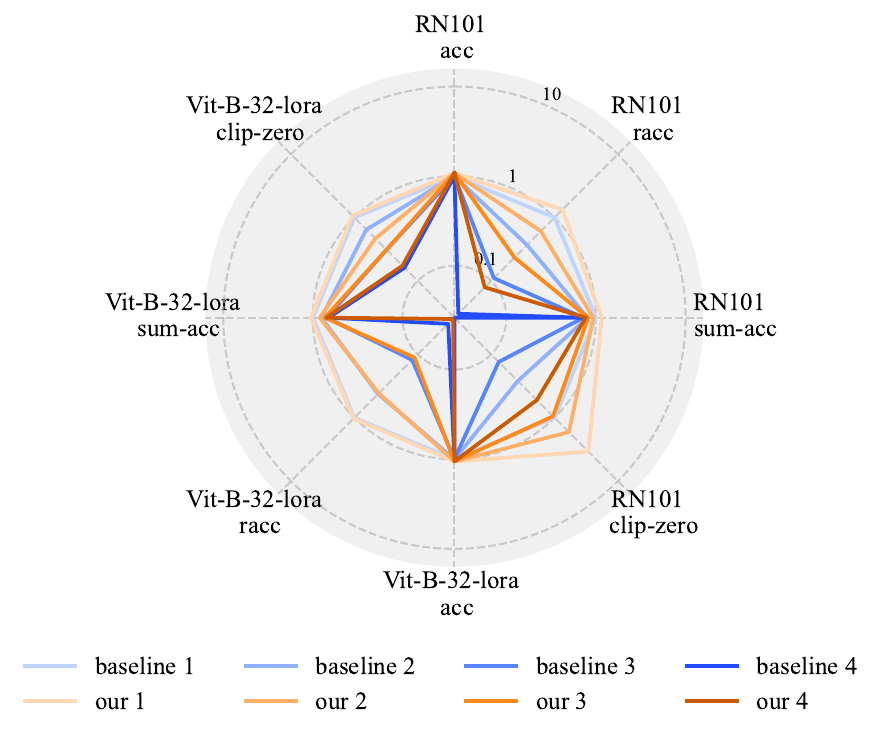}
        \caption{}
        \label{fig:sub2}
    \end{subfigure}
    \caption{ Multidimensional performance comparison of MMT-ARD with the baseline under different backbone. (a) Teacher-student combination based on ViT-B-32 and RN50. (b) Combination based on ViT-B-32-lora and RN101. The method proposed in this study (Our 1-4) comprehensively outperforms the baseline methods (Baseline 1-4) across the clean accuracy (acc) and robust accuracy (racc).}
    \label{fig:radar}
\end{figure}
To break through th
Current mainstream defenses fall into three categories: adversarial training, parameter-efficient fine-tuning, and knowledge distillation. Adversarial training enhances robustness by minimizing adversarial loss, but is computationally expensive\cite{011,013,014}. Parameter-efficient fine-tuning methods (e.g., prompt tuning) reduce computational requirements yet rely heavily on the inherent robustness of pre-trained models, leading to poor cross-dataset generalization\cite{008,009}. Knowledge distillation, particularly Adversarial Robustness Distillation (ARD), have shown great potential in enhancing model resilience. However, existing approaches still suffer from three key limitations: 1) Foundational fine-tuning flaw: they rely on fine-tuning non-robust large models as teachers, which is costly and ineffective in addressing the inherent structural vulnerabilities; 2) Convergence efficiency bottleneck: Student models require hundreds of epochs to approach teacher performance, making it difficult to meet practical deployment efficiency requirements; and 3) Single-teacher architecture limitation: A single teacher cannot simultaneously transfer both strong discriminative (clean) and robust (adversarial) features, resulting in an inevitable trade-off between clean accuracy and robustness.
ese limitations, we propose a Multimodal Multi-Teacher Adversarial Robustness Distillation (MMT-ARD) framework. The main contributions are summarized as follows:
1. A \textbf{multimodal multi-teacher knowledge fusion architecture} is designed to achieve synergistic optimization between clean feature preservation and robust feature enhancement.
2. A \textbf{Dynamic Importance Weighting (DIW) algorithm} is proposed to adaptively balance the knowledge transfer intensity from multiple teachers based on confidence and feature relevance.
3. A \textbf{cross-modal consistency constraint loss} is constructed to enhance adversarial invariance within the visual-textual embedding space, improving the model’s robustness under multimodal perturbations.

Extensive experiments on  ImageNet and Zero-Shot  benchmarks demonstrate the effectiveness of the proposed method, showing significantly improvements on ViT-B-32 robustness by 4.32$\%$, zero-shot accuracy by 3.5$\%$, and training efficiency by 2.3x over traditional adversarial distillation approaches.
As shown in Figure \ref{fig:radar}. It can be clearly observed that under different architectures (such as ResNet and Vision Transformer), the performance polygon of our method ('Our') significantly encloses that of the baseline ('Baseline'), indicating that our method achieves overall performance improvements in clean accuracy, robust accuracy, and generalization metrics.

\section{Related Work}

\subsection{Adversarial  Attack}

Adversarial attacks aim to mislead deep learning model by adding carefully crafted perturbations. Depending on the attacker’s level of knowledge about the target model, adversarial attack research has evolved into three categories: optimization-driven attacks (e.g., FGSM~\cite{000}, PGD~\cite{001}) which iteratively optimize perturbations to maximize prediction errors; attention-reconstruction attacks (e.g.,AOA~\cite{002}, TAIG~\cite{003}) which manipulate the model’s attention maps to disrupt feature localization; and decision-smoothing attacks (e.g.,TI~\cite{004}, DI~\cite{005}) which improve transferability by smoothing the loss. Hybrid methods such as SM2I-FGSM~\cite{006}  combine these strategies to exceed the limits of single-mechanism attacks. With the popularity of multimodal foundation models, adversarial research has expanded toward attacking multi-model cooperative systems.

\subsection{Adversarial Robustness via Finetuning}
Traditional single-modal defenses such as SAT \cite{007} and TRADES \cite{021} improve robustness through min-max optimization but fail under cross-modal attacks \cite{017} and suffer significant drops in zero-shot generalization performance \cite{019}, which limits their utility in open environments. In contrast, multimodal cooperative defense offers a more systematic and resilient solution. Text-guided contrastive defenses (e.g., PMG-AFT) improve robustness by freezing the text encoder to stabilize the shared feature space \cite{020}, thus achieving robust accuracy gains on ImageNet.  Meanwhile, cross-modal feature alignment methods (e.g., FARE) employ unsupervised adversarial fine-tuning, which eventually reduces the adversarial feature bias to below 0.1. More importantly, multimodal defense establishes a "cross-modal immune system" \cite{016}, which greatly improves the defense rate of joint attacks in scenarios such as payment systems. Collectively, these advances demonstrate that vision-language joint optimization effectively overcomes the cross-modal vulnerability of single-modal defense and provides a robust and generalizable protection mechanism for open environments.

\subsection{Knowledge Distillation}
The core framework of knowledge distillation~\cite{yang2024clip, yang2022cross, yang2021hierarchical} is to transfer valuable knowledge from the teacher to the student. Traditional Robust knowledge Distillation methods (such as RSLAD \cite{022}) introduce robust soft labels but  remains constrained by the single-teacher ceiling: student performance cannot surpass that of its teacher. The defense success rate under black-box attacks is still less than 50$\%$. Traditional single-teacher adversarial robust distillation exposes the modal fragmentation predicament: visual teachers cannot guide text adversarial defense, resulting in fatal vulnerabilities in multimodal system defense.
 The multi-teacher knowledge distillation framework introduced to the study of adversarial distillation \cite{018,023}. It is worth noting that our research extends multi-teacher distillation to both robust and multimodal large language model contexts. The key intuition is that different robust teacher models (trained via distinct adversarial strategies) possess complementary strengths in handling various input regions or semantic attributes\cite{026,028,029}. By allowing the student to learn collaboratively from multiple robust teachers, the proposed framework enables the integration of diverse robustness cues, producing student models that not only inherit but often surpass the robustness of any individual teacher.

\section{Method}

\subsection{Multimodel Multi-Teacher Adversarial Robust Distillation}
\begin{figure*}[htbp]
    \centering
    \includegraphics[width=1\textwidth]{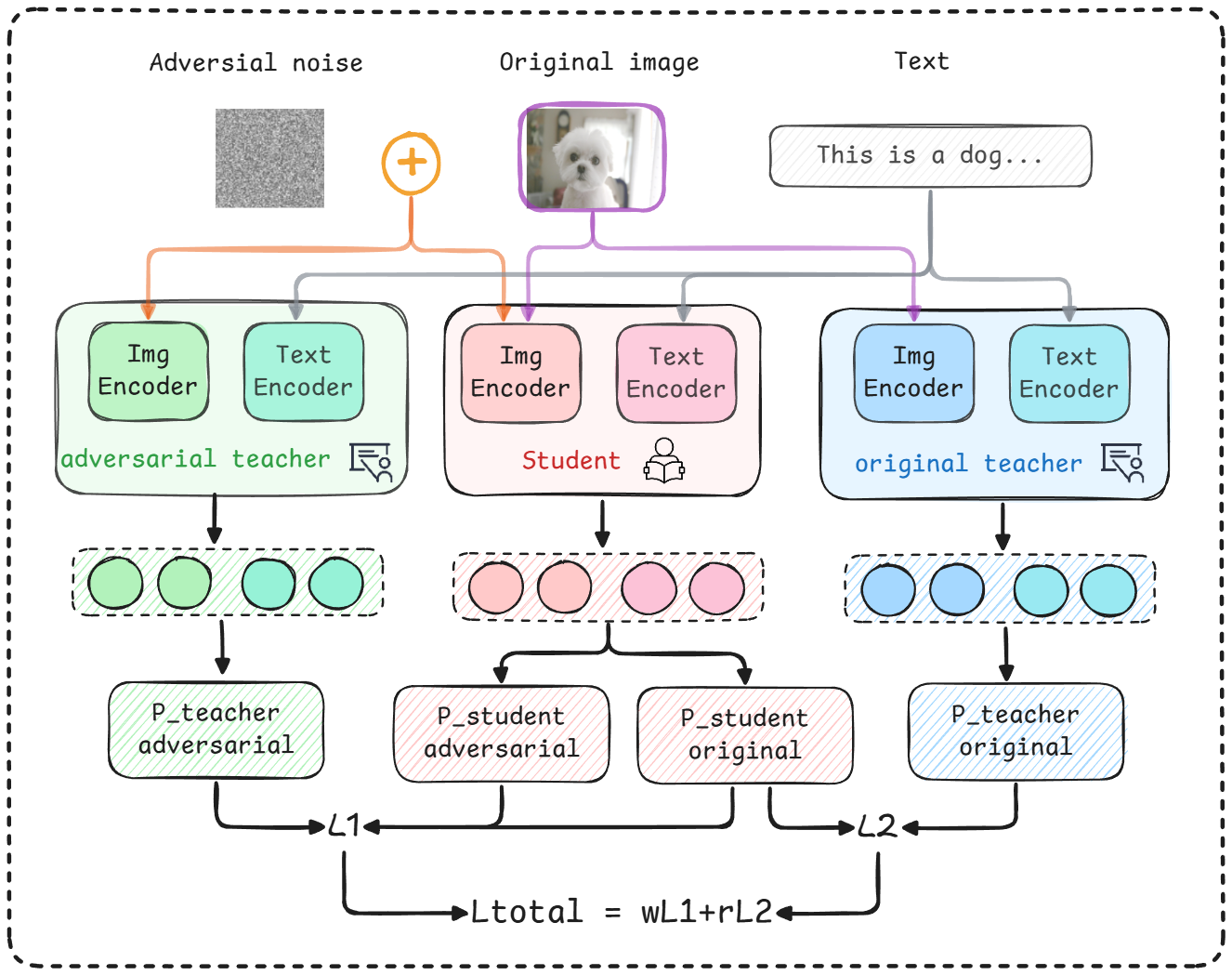} 
    \caption{MMT-ARD framework architecture, where the same input image is processed separately by two sets of encoders from the original teacher and the adversarial teacher. L1 and L2, which respectively constrain the consistency of the student model's outputs with those of the two teachers, ultimately achieving collaborative transfer of robust representations through a weighted sum.}
    \label{fig:total}
\end{figure*}

Inspired by multi-teacher and robust unsupervised finetuning, we propose the Multimodel Multi-Teacher Adversarial Robust Distillation (MMT-ARD) framework. The core idea of this method is to simultaneously utilize an Adversarial Teacher and a Clean Teacher to guide the training of a student CLIP model, thereby significantly improving the robustness of the model under adversarial attacks while maintaining the consistency of its multimodal embeddings. This design ensures consistent cross-modal feature representations while maintaining strong performance under both clean and adversarial conditions. The overall architecture of our proposed MMT-ARD framework is illustrated in Figure \ref{fig:total}.

We employ the fine-tuned CLIP model as the adversarial teacher and the original CLIP model as the clean teacher. During training, the student model is jointly supervised by both teachers: the adversarial teacher provides a robust feature representation under adversarial samples, whose input is the adversarial samples generated when the student model is internally maximized, while the clean teacher provides a semantic feature representation under clean samples. The student model receives both adversarial and clean inputs, producing outputs that are guided by corresponding adversarial soft labels and clean soft labels. Therefore, the robustness optimization framework of the proposed MMT-ARD method can be formulated as follows:
\begin{align}
O_{\mathrm{FT}} = \arg\min \sum_{i=1}^n \Big[
    &(1-\alpha)\cdot KL\left(S_{\mathrm{org}}(x_i),T_{\mathrm{org}}(x_i)\right)  \nonumber \\
    &+ \alpha\cdot KL\left(P_S(x_i),P_T(x_i)\right)
\Big],
\end{align}


\begin{equation}P_m(x)=\max_{\delta\leq\varepsilon}\left\|m_{adv}(x+\delta)-m_{\mathrm{org}}(x)\right\|_2^2,\end{equation}

\noindent
where $\delta$ defines the perturbation constraint for generating adversarial samples, ensuring that the resulting perturbations are imperceptible to the human eye. Specifically, this constraint limits the pixel-wise change in an image to not exceed a small positive threshold $\epsilon$, thereby preserving the visual appearance of the original input. Among them, $S_\mathrm{org}$ represents the clean student model, $T_\mathrm{org}$ represents the clean teacher model, $m_\mathrm{adv}$ represents the adversarial student model or the adversarial teacher model, and $max$ represents the element with the largest absolute value within the feature space. The hyperparameter $\alpha$ controls the relative importance of the two sub-objectives in the final optimization process. By adjusting $\alpha$, the training process can flexibly balance the emphasis between clean and adversarial objectives. In the following section, we introduce an adaptive parameter mechanism designed to dynamically regulate this weighting within the loss function.

\subsection{Dynamic Weight of Teachers' Confidence}
In the multi-teacher distillation framework, traditional static weight distribution methods exhibit two key limitations. First, the reliability of knowledge sources differs inherently between teachers. The adversarial teacher’s prediction confidence for adversarial samples typically shows a bimodal distribution—where high-confidence correctly defended samples coexist with low-confidence attacked samples—whereas the clean teacher’s confidence distribution on original samples is unimodal and more stable. Second, the weight distribution should have sample dependence: predictions for simple categories (e.g., “dog”) tend to be more confident than those for complex scenes (e.g., “crowded marketplace”). Static weighting, therefore, cannot adapt to the semantic complexity and difficulty of different samples. To address these issues, we propose a dynamic weight allocation strategy grounded in three core principles: 1) Deterministic priority principle: Assign higher weights to high-confidence predictions to ensure reliable knowledge transfer. 2) Uncertainty penalty principle: Suppress the interference of noise signals by reducing the weight of low-confidence predictions. 3) Cross-modal alignment principle: Promote the consistency of multimodal representations through the joint estimation of confidence across visual and linguistic modalities. This dynamic weight distribution strategy essentially builds a sample-adaptive knowledge fusion mechanism, enabling the model to automatically adjust the degree of trust assigned to different teachers based on specific sample features, thereby achieving more precise and robust knowledge distillation.

\textit{Definition of Teacher Confidence}: Given a teacher model $T$ and an input $x$, its prediction confidence is:

\begin{equation}\operatorname{conf}_T(x)=\max(\sigma(T(x))),\end{equation} where $\sigma()$ denotes the softmax function and $T(x)\in\mathbb{R}^C$ is the categorical logits vector. Dynamic weight calculation: for adversarial teacher $T_\mathrm{adv}$ and clean teacher $T_\mathrm{org}$, define the weight ratio as follows.

\begin{equation}\rho(x)=\frac{\mathrm{conf}_{T_{adv}}(x)}{\mathrm{conf}_{T_{org}}(x)+\upsilon},\end{equation} where $\upsilon=10^{-5}$ is the numerically stable term. The final weights are generated using the modified sigmoid function.

\begin{equation}w_{adv}(x)=\frac{1}{1+e^{-\lambda(\rho(x)-\tau)}},\end{equation}

\begin{equation}w_{clean}(x)=1-w_{adv}(x),\end{equation} where $\lambda$ denotes the slope coefficient, which controls the sharpness of the weight change and 
$\tau$ is the offset to adjust the weight balance.

\section{Theoretical Analyses}

\textbf{Robustness Transfer under Multi-Teacher Distillation.}
Let $\{z^{(m)}:\mathcal{X}\to\mathbb{R}^K\}_{m=1}^M$ be $M$ teacher logit maps with nonnegative weights $w_1,\ldots,w_M$ such that $\sum_m w_m=1$. For a labeled input $(x,y)$, define the \emph{teacher margins} $\gamma^{(m)}(x):=z^{(m)}_y(x)-\max_{k\neq y} z^{(m)}_k(x), m=1,\dots,M$, and the \emph{logit-averaged ensemble} $z^{\mathrm{ens}}(x):=\sum_{m=1}^M w_m\, z^{(m)}(x)$, $\Gamma_{\mathrm{ens}}(x):=z^{\mathrm{ens}}_y(x)-\max_{k\neq y} z^{\mathrm{ens}}_k(x)$. Suppose that the student logit map $z^S:\mathcal{X}\to\mathbb{R}^K$ is $L_S$–Lipschitz w.r.t.\ $\ell_2$ and fits the ensemble at $x$ within $\ell_\infty$ discrepancy $\Delta(x):=\bigl\|z^S(x)-z^{\mathrm{ens}}(x)\bigr\|_\infty$. Then for any perturbation $\delta$ with $\|\delta\|_2\le\varepsilon$, the student’s perturbed margin satisfies the 
following:
\begin{align}
\underbrace{\,z^S_y(x+\delta)-\max_{k\neq y} z^S_k(x+\delta)\,}_{\text{student margin at }x+\delta}
\;\;\ge\;\;& 
\underbrace{\sum_{m=1}^M w_m\,\gamma^{(m)}(x)}_{\text{avg.\ teacher margin at }x} \nonumber\\
& -\;2\Delta(x)\;-\;2L_S\,\varepsilon.
\end{align}
In particular, the student’s top-1 prediction at $x+\delta$ remains $y$ for all $\|\delta\|_2\le\varepsilon$ whenever
$$
\varepsilon\;<\;\frac{\sum_{m=1}^M w_m\,\gamma^{(m)}(x)-2\Delta(x)}{2L_S}.
$$

\noindent
\textbf{Ensemble margin \textit{vs.} average teacher margins.}
By convexity of $\max$, for any vectors $a^{(m)}\in\mathbb{R}^K$,
$\max_k\big(\sum_m w_m a^{(m)}_k\big)\!\le\!\sum_m w_m \max_k a^{(m)}_k$.
With $a^{(m)}\!=\!z^{(m)}(x)$, we get $\Gamma_{\mathrm{ens}}(x)\!\ge\!\sum_m w_m \gamma^{(m)}(x).$

\noindent
\textbf{Student–ensemble closeness.}
From $\|z^S(x)-z^{\mathrm{ens}}(x)\|_\infty\le\Delta(x)$, $z^S_y(x)\ge z^{\mathrm{ens}}_y(x)-\Delta(x)$ and $\max_{k\neq y} z^S_k(x)\le \max_{k\neq y} z^{\mathrm{ens}}_k(x)+\Delta(x)$, we have $z^S_y(x)-\max_{k\neq y} z^S_k(x)\;\ge\;\Gamma_{\mathrm{ens}}(x)-2\Delta(x) \stackrel{(1)}{\ge}\sum_m w_m \gamma^{(m)}(x)-2\Delta(x).$

\noindent
\textbf{Lipschitz stability.}
Since each logit of $z^S$ is $L_S$–Lipschitz,
$|z^S_c(x+\delta)-z^S_c(x)|\le L_S\|\delta\|_2$ for all classes $c$.
Thus the margin can shrink by at most $2L_S\|\delta\|_2$:
$$
z^S_y(x+\delta)-\max_{k\neq y} z^S_k(x+\delta)
\;\ge\;\bigl(z^S_y(x)-\max_{k\neq y} z^S_k(x)\bigr)-2L_S\|\delta\|_2.
$$
Combining Eqs. (2) and (3) gives the claim; the robustness condition follows by positivity of the right-hand side.

\begin{remark}
The theorem states the student inherits robustness from multiple teachers through their average margin, but loses some due to imperfect matching of the ensemble and sensitivity to input changes. To strengthen guarantees, increase teachers’ margins, reduce the student–ensemble mismatch during distillation, and control the student’s Lipschitz constant.
\end{remark}
\section{Experiment}

\subsection{Dataset Description}
ImageNet-1K\cite{024} serves as the main primary dataset for both training and evaluation, where adversarial samples are generated to assess model robustness. Additionally, we evaluate the model’s generalization capability on zero-shot classification tasks following the standard zero-shot evaluation protocol of the CLIP pre-trained model.

\subsection{Implementation Details}

\textbf{Model architecture and training configuration:} 
For model selection, we used OpenFlamingo 9 B and LLaVA-1.5 7 B as LVLM models as the infrastructure for teacher and student models. For the Teacher model, we choose dual teachers (adversarial teacher and Clean teacher), including ViT-L-14\_PMG\_Fast2 (adversarial training version and self-trained Clean Teacher (based on ViT-L-14). On the student model, we experiment with four networks, including ViT-B-32, RN50, RN101, ViT-B-32-LoRA (using the LoRA fine-tuning strategy). All experiments were performed under the same hardware environment (NVIDIA A100), and the results were repeated three times and averaged to ensure statistical significance.

\noindent
\textbf{Evaluation metrics include:} Clean Accuracy (acc) : The classification accuracy of the model on clean samples.
Robust Accuracy (racc): The classification accuracy of the model on adversarial samples. Adversarial samples are generated using PGD attacks, with attack intensities ($\epsilon$) of 1/255, 2/255, 3/255, and 4/255, respectively.
Sum-ACC: The Sum of clean accuracy and robust accuracy, which is used to comprehensively evaluate the model performance.
Zero-Shot Accuracy: Accuracy on zero-shot classification tasks.

\subsection{Comprehensive Comparative Experiments on MM-TARD}

\begin{table*}[!ht]
    \centering
    \tabcolsep 0.1cm
    \caption{Performance of the benchmark method and the proposed method under the MMT-ARD framework on ViT-B-32, ResNet-50, ResNet-101 and ViT-B-32-Lora models}
    \resizebox{\linewidth}{!}{
    \begin{tabular}{ccccc ccccc ccccc ccc} 
    \hline
    \multirow{2}{*}{Method} & 
\multirow{2}{*}{eps}
    &\multicolumn{4}{c}{CLIP ViT-B-32} & 
    \multicolumn{4}{c}{CLIP RN50} & \multicolumn{4}{c}{CLIP RN101} & 
        \multicolumn{4}{c}{CLIP ViT-B-32-Lora}\\
        \cmidrule(lr){3-6} \cmidrule(lr){7-10} \cmidrule(lr){11-14} \cmidrule(lr){15-18}
        ~ &  & acc & racc & sum-acc & clip-zero & acc & racc & sum-acc & clip-zero  & acc & racc & sum-acc & clip-zero & acc & racc & sum-acc & clip-zero  \\ \hline
        \multirow{4}{*}{baseline} & 1 & 61.84 & 49.00 & 110.84 & 26.40 &  43.92 & 23.92 & 67.84 & 6.5 & 45.84 & 20.44 & 66.28 & 3.8 &  43.24 & 22.46 & 65.70 & 16.10  \\ 
        ~ & 2 & 61.84 & 34.56 & 96.40 & 19.20  & 43.92& 10.14 & 54.06 & 3.1 & 45.84 & 7.54 & 53.38 & 1.0 & 43.24& 9.46 & 52.70 & 10.40   \\ 
        ~ & 3 & 61.84 & 22.72 & 84.56 & 14.00 &  43.92 & 4.04 & 47.96 & 1.8 & 45.84 & 2.26 & 48.1 & 0.5 & 43.24& 2.74 & 45.98 & 5.0   \\ 
        ~ & 4 & 61.84 & 13.76 & 75.56 & 9.90 &  43.92 & 1.28 & 45.2 & 1.0  & 45.84 & 0.62 & 46.46 & 0.1 & 43.24& 0.74 & 43.98 & 2.6   \\ \hline
        \multirow{4}{*}{our} & 1 & 63.48 & 49.34 & 112.82 & 27.10 &  46.56 & 25.36 & 71.92 & 9.0 & 49.48 & 27.30 & 76.78 & 13.0 &  45.76 & 23.06 & 68.82 & 17.2  \\ 
        ~ & 2 & 63.48 & 34.78& 98.26 & 19.60 &  46.56 & 10.94 & 57.5 & 5.1 & 49.48 & 12.46 & 61.94 & 6.4 &  45.76 & 9.24 & 55.0 & 7.5 \\ 
        ~ & 3 & 63.48 & 22.24 & 85.72 & 13.80 &  46.56 & 4.28 & 50.84 & 3.0  & 49.48 & 4.78 & 54.26 & 3.6 &  45.76 & 2.52 & 48.28 & 5.0 \\ 
        ~ & 4 & 63.48 & 12.92 & 76.42 & 9.60 &  46.56 & 1.46 & 48.02 & 1.6 & 49.48 & 1.62 & 51.10 & 2.0 &  45.76 & 0.62 & 46.38 & 2.8 \\ \hline
    \end{tabular}}
    \label{tabel:total}
\end{table*}

\subsubsection{Enhanced robustness}
As shown in Table \ref{tabel:total}, under low-intensity attack scenarios, our method achieves a 4.32$\%$ absolute improvement in robust accuracy (racc) over the baseline (from 45.02$\%$ to 49.34$\%$), representing a statistically significant enhancement. This demonstrates that the proposed dual-teacher distillation strategy effectively strengthens the model’s resilience to adversarial perturbations.  More importantly, the model also exhibits an absolute gain of 3.5$\%$ in zero-shot accuracy, indicating that that by learning more discriminative feature representations from the clean teacher, it acquires superior generalization capabilities rather than merely overfitting to adversarial examples. Furthermore, the increase in the overall Sum-acc metric (+2.48) further validates the comprehensive optimization effect of the proposed approach on the model’s robustness and generalization performance.

\subsubsection{High-intensity attack}
As the attack intensity ($\epsilon$) increases, the distribution difference between adversarial samples and clean samples intensifies, and the performance of all models declines as expected. Under this extreme setting, our method performs close to the baseline in terms of robustness, but maintains an advantage of approximately 1.6$\%$ in clean accuracy consistently. This indicates that our method does not lose robustness in extreme adversarial environments, and at the same time successfully enables the student model to learn representations that are closer to the essential features of natural images, thereby achieving better performance on clean data.

\subsubsection{Generalization verification}
The results on the ResNet architecture further verify the universality of our method. For RN101, our method achieves absolute improvements of 3.64$\%$ in clean accuracy and 2.52$\%$ in robust accuracy. Most importantly, its robust accuracy has more than doubled (a relative increase of 111.5$\%$), while the zero-shot performance has improved by 3.1$\%$ (a relative increase of 720$\%$). These results demonstrate that the proposed dual-teacher distillation strategy is effective across models with varying capacities and architectures. In particular, it substantially enhances the adversarial robustness of classical architectures like ResNet while preserving strong transferability and generalization performance.

\begin{figure*}[htbp]
    \centering
    
    \includegraphics[width=\textwidth]{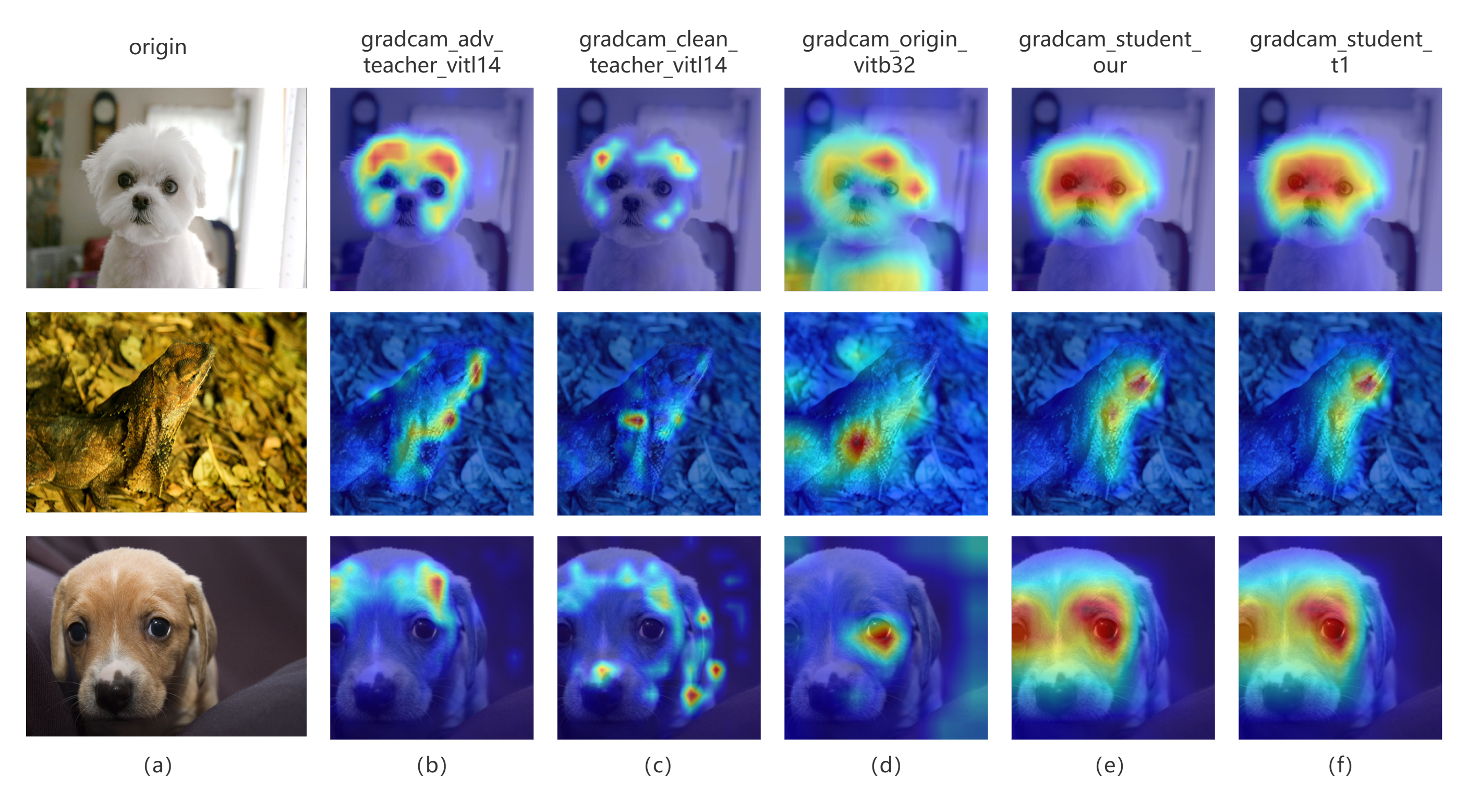} 
    
    \vspace{-3mm} 
    
    
    \caption{Heatmaps of the models for different teacher-student pairs.}
    \label{fig:heatmap}
\end{figure*}

Taking the above analysis together, our multi-teacher distillation method can work effectively on multiple architectures such as ViT and ResNet, and its core advantages are reflected in: 1) significantly improving the robustness and generalization ability of the model under common low-intensity attacks; 2) Maintain competitiveness under high-intensity attacks and optimize the essential feature representation of the model; 3) It shows excellent generalization for different model architectures; 4) Perfect compatibility with efficient parameter fine-tuning technology, with high practical value. Figure \ref{fig:heatmap}shows the visualization of the experimental results.
Figure \ref{fig:heatmap}. (a) represents the original, clean input image,(b) represents the Grad-CAM heat map generated by the adversarial teacher (ViT-L-14) when processing the adversarial examples,(c) represents the Grad-CAM heat map generated by the clean teacher (ViT-L-14) when processing the clean original image, and (c) represents the Grad-CAM heat map generated by the clean teacher (Vit -L-14) when processing the clean original image. Heat map of (d) the original student model without distillation (ViT-B-32) on the original image,(e) the student model distilled by our proposed multi-teacher method on the original image, and (f) the student model distilled using only a single teacher (adversarial teacher). The figure clearly reveals different models (teacher vs. student) and different methods (baseline vs. student). Our approach) fundamental differences in the basis for decision making.

\subsection{Ablation Study}
To comprehensively analyze the performance of our proposed MMT-ARD framework and verify the effectiveness of the contribution of each component, we conducted systematic ablation studies. This section addresses three core questions: (1) What improvements are brought by introducing the clean teacher and its integration strategy? (2) How do different loss function designs affect the trade-off between accuracy and robustness of the model? (3) How should the supervisory signals from multiple teachers be balanced to achieve optimal performance? We explore these aspects through controlled experiments, isolating the effect of each factor.
\subsubsection{Path-separated dual teacher strategy}
This experiment evaluates the necessity of introducing a clean teacher and a confidence-based weighting strategy. We compared three strategies: 1. Baseline: Uses only the adversarial teacher model (ViT-L-14 PMG Fast2) with the student model ViT-B-32. 2. Average: Uses both the adversarial and clean teachers; their output embeddings are averaged with equal weights. 3. Path-Separated Dual Teachers \textbf{(Ours)}: Employs both teachers, where their predictions are dynamically weighted and fused based on confidence levels. 

As shown in Table 2, introducing a clean teacher consistently improves performance. Compared with the baseline, the equal-weight averaging strategy achieves minor improvements of +0.26$\%$ in clean accuracy and +0.16$\%$ in robust accuracy, with a Sum-acc increase of 0.42$\%$. This demonstrates that discriminative features learned from the clean teacher (derived from natural image distributions) complement the robust features of the adversarial teacher. 

However, our path-separated dual-teacher strategy further enhances performance.  While maintaining robust accuracy (racc: 34.72$\%$), the clean accuracy improves by 0.38$\%$ over the baseline. This indicates that allowing the clean teacher to focus on generating highly discriminative target embeddings for the original images provides a better learning target for the student model, thereby improving classification performance. Compared with the simple Average strategy: the clean accuracy further improves by 0.12$\%$, confirming that naive output fusion is suboptimal. 

\begin{table}[!ht]
    \centering
    \caption{Experimental results of different combinations of teachers. CA: Clean Accuracy, RA: Robust Accuracy, Baseline: (Adv. Teacher Only)}
    \renewcommand{\arraystretch}{1.0}
    \begin{tabular}{llll}
        \hline
        \textbf{Strategy } & \textbf{CA (acc)} & \textbf{RA (racc)} & \textbf{Sum-acc} \\ 
        \hline
        \textbf{Baseline } & 61.96  & 34.56 & 96.52 \\ 
        \textbf{Average} & 62.22  & 34.72  & 96.94  \\
        \textbf{Weighted (Ours)} & 63.48  & 34.78  & 98.26  \\ 
        \hline
    \end{tabular}
\end{table}

\subsubsection{Dynamic weighting strategy based on teachers' confidence}
This experiment compare three configurations: 1. Single-KL (Baseline): Uses only the adversarial teacher’s output as soft labels to compute the KL divergence loss. 2. Dual-KL (0.5:0.5): Computes KL divergence from both teachers with equal (1:1) weighting. 3. Dual-KL + Adaptive Norm: Extends Dual-KL by introducing an adaptive normalization loss. As shown in Table 3, this experiment clearly illustrates the accuracy–robustness trade-off. When switching from Single-KL to fixed-weight Dual-KL loss, the model learns extremely discriminative features from the clean teacher, resulting in a sharp increase in clean accuracy by +10.18$\%$. However, this aggressive optimization deviates significantly from the robust feature space guided by the adversarial teacher, causing a sharp drop in robust accuracy by -23.02$\%$. This indicates that giving equal weight to both teachers in the loss function leads to severe gradient conflicts in the optimization objective, making it difficult for the student model to simultaneously fit two highly divergent distributions. After adding the Adaptive Norm loss, the clean accuracy is further increased, but robustness is almost completely lost confirming that static fusion cannot effectively balance the competing learning signals. In contrast, incorporating a dynamic confidence-based weighting strategy  significantly improves overall performance: compared to the baseline, clean accuracy is significantly improved by +1.52$\%$, robust accuracy reaches 34.78$\%$, and Sum-acc significantly gains +1.74$\%$. 

These results demonstrate that static averaging is suboptimal, while dynamic weighting enables adaptive balancing. When the adversarial teacher exhibits high confidence, the model prioritizes its supervision to preserve robustness; when confidence is low, it relies more on the clean teacher’s discriminative features to enhance accuracy. This adaptive cooperation between teachers is key to achieving balanced and superior performance.
\begin{table}[!ht]
    \centering
    \caption{Experimental results for different loss function designs. CA: Clean Accuracy, RA: Robust Accuracy.}
    \renewcommand{\arraystretch}{1}
    \begin{tabular}{lll}
    \hline
        \textbf{Loss Design} & \textbf{CA (acc)} & \textbf{RA (racc)} \\ \hline
        \textbf{Single-KL (Baseline)} & 61.96 & 34.56 \\ 
        \textbf{Dual-KL (0.5:0.5)} & 72.14  & 11.54  \\ 
        \textbf{Dual-KL + Adaptive Norm} & 73.26 & 0.34 \\ \hline
    \end{tabular}
\end{table}

\subsubsection{Loss weight}
Based on the findings in Section 5.4.2, we conducted an in-depth analysis to optimize the accuracy–robustness trade-off by fine-tuning the loss weight ratio between the two teachers ($\lambda_{adv}$: $\lambda_{org}$). The experiment successfully found the optimal operation point (Sweet Spot). As shown in Table 4, when the weight ratio is set to 3:0.5 (i.e., $\lambda_{adv}$ / $\lambda_{org}$ = 6), the model achieves an optimal balance between clean accuracy (63.88$\%$) and robust accuracy (34.42$\%$). Both indicators at this point are significantly better than the Dual-KL (0.5:0.5) setting, and the robustness returned to a level comparable to the baseline, while the clean accuracy maintained an improvement of nearly 2$\%$. 
These results highlight three key insights: 1. Adversarial supervision should dominate the training process: a higher $\lambda_{adv}$ ratio is a prerequisite for maintaining model robustness. This is in line with the essence of adversarial training, that is, the model must prioritize learning stable decision boundaries. 2. Clean supervision refines representations: A small but non-zero $\lambda_{org}$ weight is sufficient to provide the necessary discriminative signal, effectively refining the basic feature representation learned from adversarial training, thereby improving the clean accuracy without compromising its stability. 3. Balance is feasible: Through strict weight tuning, a new Pareto Optimal point can be found, breaking through the trade-off boundary between robustness and accuracy without significantly sacrificing robustness, and achieving an overall performance improvement.
\begin{table}[!ht]
    \centering
    \caption{Experimental results of different loss weight proportions.CA: Clean Accuracy, RA: Robust Accuracy}
    \renewcommand{\arraystretch}{1}
    \begin{tabular}{lll}
    \hline
        \textbf{Weight Ratio $\lambda_{adv}$ :$\lambda_{org}$} & \textbf{CA (acc) } & \textbf{RA (racc)} \\ \hline
        \textbf{1 : 0.5} & 71.26 & 16.18  \\ 
        \textbf{2 : 0.5} & 68.72  & 21.84 \\ 
        \textbf{3 : 0.5} & 63.88 & 34.42 \\ 
        \textbf{3.5 : 0.5} & 63.74 & 34.44  \\ 
        \textbf{3 : 1 } & 69.88 & 18.76  \\ 
        \textbf{7 : 0.3 } & 62.88 & 34.60 \\ \hline
    \end{tabular}
\end{table}

\subsubsection{Quantitative analysis of gradient}
To quantitatively evaluate the consistency of visual attention regions between different distillation models and their teacher model, we adopt a method based on Grad-CAM feature map subtraction followed by $L2$ norm computation. The numerical value intuitively reflects the degree of difference in the attention region between the models. The resulting L2 norm intuitively reflects the degree of discrepancy between the attention regions: a smaller value indicates greater similarity between the student’s  and teacher’s gradient feature maps, implying stronger alignment with the teacher’s guidance. 

As shown in Table 5, analyses conducted on three validation set images (ILSVRC2012\_val\_00004748, ILSVRC2012\_val\_00012820, ILSVRC2012\_val\_00014409) reveal that reveal that the proposed method consistently achieves significantly lower L2 norm values than ViT-B-32, whether compared against the clean or adversarial teacher.  This strongly proves that our approach can effectively make the student model learn and inherit the key feature attention regions of the teacher model, thereby improving feature representation transfer efficiency. 

In the path distilled from adv\_teacher, our method achieves L2 values (2157, 2296, 2571) lower than or equal to the baseline method (2175, 2316, 2580) on all three images, indicating a slight but consistent advantage in capturing the attention mechanism of the robust teacher model. It reflects the positive effect of the introduced module. Therefore, from the perspective of gradient feature similarity, this experiment confirms that the multi-teacher distillation framework proposed in this paper can effectively promote the student model to align the visual attention of the teacher model more accurately, thus ensuring the effectiveness of knowledge distillation at the feature level, which lays a foundation for the performance improvement of the final model.
\begin{table}[!ht]
    \centering
    \caption{Comparison of the knowledge distillation effectiveness of Clean teacher (Cle\_T) and adversarial teacher (adv\_T) models for ViT-B-32, baseline, and MMT-ARD (quantitative results on datasets of ILSVRC2012\_va1\_00004748 (Val\_1), ILSVRC2012\_val\_00012820 (Val\_2), and ILSVRC2012\_val\_00014409 (Val\_3) respectively.)}
    \renewcommand{\arraystretch}{1}
    \begin{tabular}{llll}
    \hline
        & \textbf Val\_1 & \textbf Val\_2 & \textbf Val\_3 \\ \hline
            \textbf{Cle\_T to ViT-B-32} & 2613 & 2603 & 2655  \\ 
        \textbf{Cle\_T to Baseline} & 2104  & 2288 &2630  \\ 
        \textbf{Cle\_T to ours} & 2107 & 2266 &2598\\ 
        \textbf{adv\_T to ViT-B-32} & 2650 & 2621 &2779 \\ 
        \textbf{adv\_T to Baseline} & 2175 & 2316 &2580  \\ 
        \textbf{adv\_T to ours} & 2157& 2296 &2571 \\ \hline
    \end{tabular}
\end{table}

\section{Conclusion}
This study have proposed a Multimodal Multi-teacher adversarial Robust distillation framework (MMT-ARD), which effectively solves the robustness problem of visual language models in adversarial environments through a dual-teacher knowledge fusion architecture and a dynamic weight allocation strategy. Experiments demonstrated that the proposed method improves robust accuracy of ViT-B-32 model by 4.32$\%$ and zero-shot accuracy by 3.5$\%$ on the ImageNet dataset, and improves the training efficiency by 2.3 times. The results of this study provide new ideas and methods for the research on the adversarial robustness of multimodal models, and provide reliable technical support for artificial intelligence applications in safety-critical fields. Future work will focus on further optimizing the dynamic weight algorithm and extending the framework to more modalities and more complex application scenarios.

{
    \small
    \bibliographystyle{ieeenat_fullname}
    \bibliography{arxiv}
}


\end{document}